\documentclass[review]{elsarticle}
\usepackage{natbib}
\biboptions{authoryear}
\usepackage{float}
\usepackage{setspace}
\usepackage{makecell}
\usepackage{booktabs}
\usepackage{xcolor}
\usepackage{csvsimple}
\usepackage{soul}
\usepackage{multirow}
\usepackage{enumitem}
\usepackage{xurl}
\usepackage{chngcntr}
\usepackage[utf8]{inputenc}
\usepackage{ragged2e}
\usepackage{graphicx}
\usepackage{amsmath}
\usepackage[section]{placeins} %force image in every section end
\sethlcolor{yellow}
\usepackage{fancyhdr}

\newcommand{\setupheaders}{
    \fancyhf{} % Clear all default fields
    \fancyhead[R]{\thepage} 
    \fancyhead[L]{\textcolor{gray!100}{\small Vanunu et al., Gradient boosting algorithm for GEO fire detection}}
    \renewcommand{\headrulewidth}{0pt} % Remove the header line
}

\setupheaders

\fancypagestyle{plain}{
    \setupheaders
}
\title{Comparing a gradient boosting algorithm to the GOES FDC for wildfire detection}

\author[1,2,3]{Asaf Vanunu}
\author[4]{Boaz Nadler}
\author[2,3]{Arnon Karnieli\corref{cor1}}
\cortext[cor1]{Corresponding author: karnieli@bgu.ac.il}

\address[1]{The Albert Katz International School for Desert Studies, The Jacob Blaustein Institutes for Desert Research, Sde Boker Campus, Ben-Gurion University of the Negev, Midreshet Ben-Gurion 84990, Israel}
\address[2]{The Goldman Sonnenfeldt School of Sustainability and Climate Change, Ben-Gurion University of the Negev, Beer Sheva 8410501, Israel}
\address[3]{The Remote Sensing Laboratory, French Associates Institute for Agriculture and Biotechnology of Drylands, The Jacob Blaustein Institutes for Desert Research, Sde Boker Campus, Ben-Gurion University of the Negev, 84990, Israel}
\address[4]{Department of Computer Science and Applied Mathematics, Weizmann Institute of Science, Rehovot 7610001, Israel}

\begin{document}
%\linenumbers
% Abstract
\begin{abstract}
Wildfires pose severe risks to human life, ecosystems, and property.
This study presents a machine learning approach for wildfire detection from GOES ABI imagery.
A CatBoost model was trained on a large dataset with thousands of ABI images and over 300,000 matching VIIRS fire detections.
An evaluation on a separate dataset across five regions showed that the learned CatBoost model outperformed the operational GOES Fire Detection and Characterization (FDC) product. It achieved higher precision, recall, and F1 scores both within and outside the training area. The CatBoost model achieved F1 scores that were 0.16 to 0.38 higher than the GOES FDC in all regions. In addition, out of 51 historical fire events, the CatBoost detected 26 fires before both VIIRS and GOES FDC, compared to only six earlier detections by the GOES FDC. Importantly, the CatBoost model achieved accurate wildfire detection also during nighttime, whereas the GOES FDC obtained very low recall values, around ~0.03. This study demonstrates that machine learning models may offer significant improvements over existing geostationary fire products, including higher accuracy, fewer false alarms, and earlier detection.
\end{abstract}
\maketitle
% Keywords
\newpage

\section{Introduction}
\label{sec:introduction}
Wildfires can have devastating impacts on the environment and pose serious threats to human life, ecosystems, and property.
A recent example is the Palisades Fire, which erupted in January 2025 in Los Angeles, California, resulting in 12 fatalities, destroyed more than 6,800 structures, and burning 95 km\textsuperscript{2} of land over 24 days before being fully contained \citep{palisades2025}.
Given the rapid spread of such severe events, timely fire detection can help contain fires and prevent large-scale disasters. This is even more crucial nowadays, as the frequency of major disasters caused by wildfires has quadrupled from 1980 to 2023 \citep{Cunningham_2025}.

A common approach to early wildfire detection is the use of satellite imagery, which provides wide-area coverage and frequent monitoring ~\citep{Chuvieco_2020_review}.
Among satellite-based methods, geostationary (GEO) satellites are commonly used for wildfire detection \citep{Barmpoutis_202_review}.
Their orbital period matches Earth’s rotation, so they appear stationary from a fixed point on the ground and can monitor the same region every 5 to 15 minutes.

Indeed, several dedicated fire detection products were developed for GEO satellite imagery.
Notable examples include the Fire Detection and Characterization (FDC) product of the American Geostationary Operational Environmental Satellite (GOES) \citep{SCHMIDT_2020_FDC}, and the Fire Radiative Power (FRP-PIXEL) product of the European Meteosat Second Generation (MSG) satellite \citep{Wooster_2015_FRP}.
These products are based on a combination of threshold and contextual criteria and have several known limitations. As discussed by \citet{SCHMIDT_2020_FDC}, some fire events are visible in the 3.7~µm band before the FDC algorithm detects them.
In addition, \citet{HALL_2019_GOES_SEVIRI, Hall_2023_GOES_H8} report that the FDC and FRP fire products exhibit high false-alarm rates.
These findings highlight the need for fire detection algorithms with lower false-alarm rates and greater detection capabilities, particularly for small fires.

A promising approach to address this challenge is to train machine learning (ML) algorithms for GEO fire detection.
ML algorithms have the potential to outperform traditional algorithms by learning fire-related complex spatial and spectral patterns from GEO satellite imagery \citep{YANG_2024_review_fire_ML_RS}.
Training a ML algorithm requires a labeled dataset consisting of multispectral imagery annotated at the pixel level as fire or non-fire.
These annotations are derived from independent reference sources.
Examples include high-resolution fire detections from Low Earth Orbit (LEO) satellite instruments \citep[e.g.,][]{Zhao_2022_DL_GOES, Zhang_2023_RF_H8_VIIRS}, and fire records from national forest services \citep[e.g.,][]{Jang_2019_ML_H8_korea, Kang_2022_DL_H8_latency, KANG_2023_Toward_H8}. After the model has been trained, given a new GEO image, its output is a binary image, indicating whether each pixel is fire or not.
Finally, it is common practice to evaluate the model's accuracy on an independent dataset that was not seen during training.

Several studies considered ML approaches for GEO fire detection.
Some works used traditional ML algorithms, such as Random Forest (RF) \citep{Jang_2019_ML_H8_korea, Zhang_2023_RF_H8_VIIRS, Maeda_2023_ML_H8_MODIS, Liu_2023_ML_H8}, while other studies applied Deep Learning (DL) algorithms, like Convolutional Neural Networks (CNNs) \citep{Hong_2022_CNN_H8, Hong_2024_SBT_DL_H8, Ding_2022_DL_H8, Zhao_2022_DL_GOES, Kang_2022_DL_H8_latency, KANG_2023_Toward_H8, Ji_2023_BRDF_DL_H8}.
These studies show the potential of ML algorithms for GEO fire detection.
Nevertheless, important limitations and gaps remain to be addressed.

One gap is that most of the published ML models for GEO fire detection used relatively small training and evaluation datasets, containing only a few hundred annotated fire pixels.
In addition, both the training and evaluation regions covered only small-scale geographic areas. Furthermore, Most works did not assess the models' generalization to regions outside the training areas.
Another issue is that, in many studies, the trained ML models were not empirically compared with the corresponding operational GEO fire detection algorithms. Furthermore, previous studies did not assess their model performance separately on daytime and nighttime images, but rather provided its overall accuracy.
Consequently, it is unclear whether these ML models truly outperform existing algorithms, particularly in terms of low false alarms and detection capability across a broad range of atmospheric conditions and over the large coverage areas of GEO satellites.

The primary goal of this study is to develop a robust ML model for GEO fire detection that outperforms the GOES FDC product in both detection accuracy and false-alarm rate.
To this end, we trained and evaluated a ML fire detection method using a large annotated dataset containing more than 60,000 GOES ABI fire pixels and $\sim$1.2 million non-fire pixels. Reference labels for both training and evaluation were derived from the LEO Visible Infrared Imaging Radiometer Suite (VIIRS) active fire product.
A Gradient Boosting ML fire detection algorithm was trained on this dataset and evaluated on a large, separate dataset. The evaluation dataset includes images from regions not used in training and from a different time period. Specifically, the training data are from 2022–2023, while the evaluation data are from 2024. These differences in both space and time between the training and evaluation datasets enable a comprehensive assessment of the model's generalization capability. In addition, the model performance was compared to the GOES FDC product in terms of overall detection and daytime/nighttime detection.

\section{Study area and data sources} 
This section describes the regions analyzed in the study and the satellite imagery acquired over them. The images were used to train and evaluate a ML model for GEO fire detection.
In addition, as detailed below, a set of specific fire events was compiled to compare the ML model's early detection capabilities with those of the GOES FDC product.    
\subsection{Study area}
\label{subsec:Study Area}

\begin{figure}[t]
    \centering
    \includegraphics[width=1\linewidth]{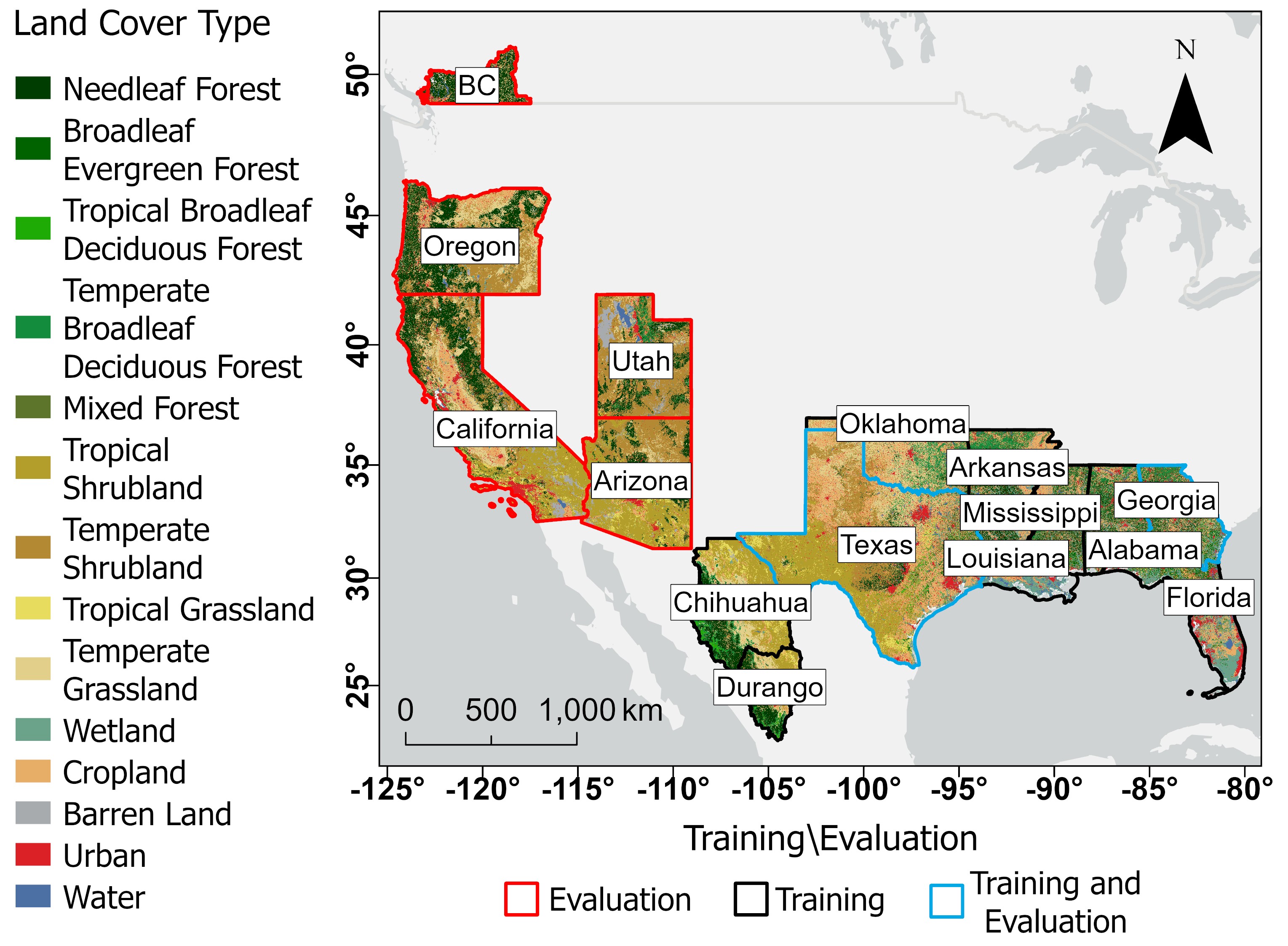}
    \caption{Study area maps. The training states are marked in black, evaluation states in red, and the states used for both are marked in light blue.}
    \label{fig: study area}
\end{figure}

The study area encompasses several regions in the United States, Mexico, and Canada, with diverse land-cover types, including forests, wetlands, deserts, agricultural and urban areas (Figure ~\ref{fig: study area}).
As illustrated in the figure, part of the study area was used solely for training, another solely for evaluation, and a third for both training and evaluation.
Specifically, the U.S. states of Florida, Alabama, Mississippi, Louisiana, and Oklahoma, as well as the Mexican states of Chihuahua and Durango, were used for training.
Texas and Georgia were used for both training and evaluation, whereas California, Oregon, Arizona, Utah, and a small region in British Columbia were used exclusively for evaluation. 
The size of the training area is approximately 2.1 million~km\textsuperscript{2}, whereas that of the evaluation region is about 2 million~km\textsuperscript{2}.

\subsection{Data sources}
\label{subsec:Data}
The training and evaluation datasets have the same structure.
Each dataset consists of VIIRS fire detections and matching GOES Advanced Baseline Imager (ABI) images acquired within a $\pm 2.5$~min interval.
Further details on the data utilized from VIIRS and GOES ABI products are provided below. 

The training dataset consists of all VIIRS fire detections in the study area throughout all of 2022 and 2023.
It contains 324,502 VIIRS fire detections and 4330 corresponding GOES ABI images.
The evaluation dataset contains VIIRS fire detections from selected months in 2024, see Table~\ref{tab:Evaluation dataset table}.
The following paragraphs elaborate on the specific data sources used to construct the training and evaluation datasets. These include the VIIRS active fire product, GOES ABI imagery, and fire perimeter data.

\begin{table}[t]
\centering
\caption{Evaluation dataset}
\vspace{0.5em}
\label{tab:Evaluation dataset table}
\resizebox{\textwidth}{!}{%
\begin{tabular}{l l c c}
\hline
\textbf{Region} & \textbf{Months (2024)} & \makecell{\textbf{GOES ABI}\\\textbf{Number of images}} & \makecell{\textbf{VIIRS}\\\textbf{Number of detected fires}} \\
\hline
Georgia          & February, March, April    & 185  & 8211  \\
Texas            & March, April, May         & 325  & 9005  \\
California-Oregon & July, August, September  & 363  & 53,259 \\
Arizona-Utah      & September, October, November & 285  & 5309  \\
British Columbia  & July, August, September      & 115  & 1523  \\
\hline
\end{tabular}%
}
\end{table}

%\paragraph{VIIRS active fire product}\mbox{}\\
\subsubsection{VIIRS active fire product}
In this work, data from the VIIRS instrument onboard the Suomi National Polar-orbiting Partnership (Suomi NPP) satellite were used as a reference.
Specifically, reference fire labels were derived from VIIRS active fire product VNP14IMG Collection 2 \citep{SCHROEDER_2014_VIIRS}.

The VNP14IMG product has a spatial resolution of 375 m. For each fire detection (pixel), it provides its location (latitude and longitude), acquisition date and time, fire radiative power (FRP), and confidence level. 
Fire detections with low confidence were deemed insufficiently reliable to serve as a reference and were thus excluded from the dataset.
The data were downloaded from NASA's Level-1 and Atmosphere Archive and Distribution System Distributed Active Archive Center (LAADS DAAC)\footnote{\url{https://ladsweb.modaps.eosdis.nasa.gov}, accessed on 1 February 2026}.

%\paragraph{GOES ABI images}\mbox{}\\
\subsubsection{GOES ABI images}
Three GOES ABI products with a spatial resolution of 2~km were downloaded: (1) Multi-channel Cloud and Moisture Imagery (MCMI); (2) Clear Sky Mask (ACM); and (3) FDC.
The MCMI and ACM data serve as input to the ML model, while the FDC data serves as the benchmark fire detector against which we compare our trained ML model.
The MCMI provides 16-band multispectral images. The complementary ACM indicates which pixels are cloudy and was used for cloud masking. The FDC product uses a contextual algorithm based on thermal infrared bands to identify thermal anomalies. It assigns each pixel a value indicating whether it is a fire. Pixels classified as fire have a corresponding confidence category. These categories reflect the algorithm's confidence level that the detected fire is indeed a fire \citep{schmidt2020goesr}.
For evaluation analysis (Section~\ref{subsec:Model evaluation}), all of the FDC confidence categories were included.
For the early detection analysis (Section~\ref{subsec:First time detection}), the evaluation was performed twice. First, only the \texttt{processed} and \texttt{saturated} categories were used, as they have the lowest false alarm rates among the FDC categories \citep{Hall_2023_GOES_H8}. Second, the \texttt{high} category was added to the previous ones. Despite its higher false-alarm rate, it was included to assess whether it enables earlier fire detection.
Both GOES-16 and GOES-18 ABI products were downloaded in Continental United States (CONUS) and Pacific U.S. (PACUS) scan modes, respectively, using the \texttt{GOES} Python library\footnote{\url{https://pypi.org/project/GOES/}, accessed on 1 February 2026} (version 3.9.23).

%\paragraph{Fire perimeter data for early detection assessment}\mbox{}\\
\subsubsection{Fire perimeter data for early detection assessment}
\begin{figure}[!ht]
    \centering
    \includegraphics[width=0.9\linewidth]{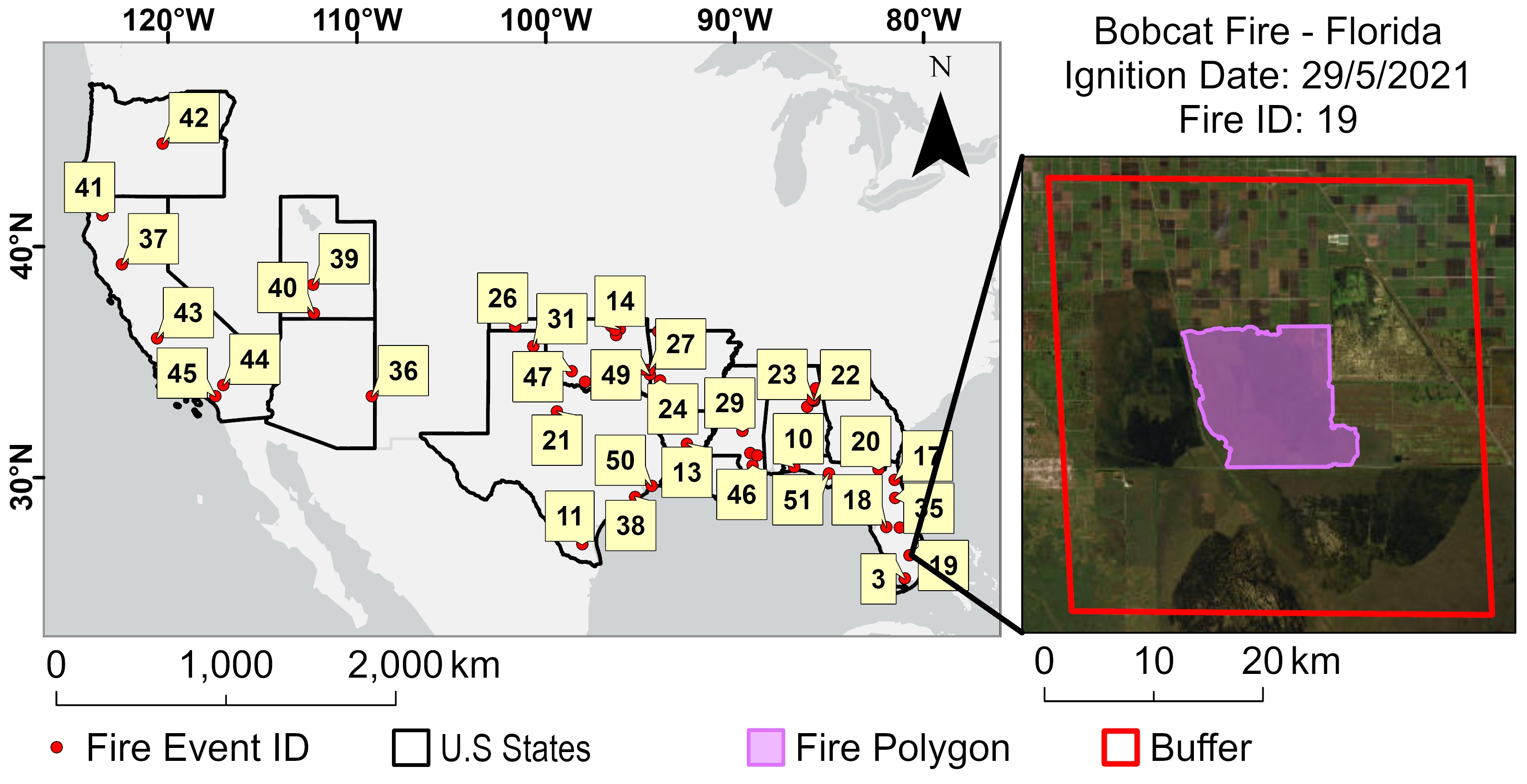}
    \caption{Map of the locations of 51 fire events with their identification numbers. For example, fire number 19 is the Bobcat Fire in Florida that occurred on 29 May 2021.}
    \label{fig:fire_perimeter_map}
\end{figure}

To compare the early-detection capabilities of the ML model and the GOES FDC (Section \ref{subsec:First time detection}), a dataset of fire-perimeter measurements for 51 specific fire events was constructed as follows. The perimeter data was retrieved from the Wildland Fire Interagency Geospatial Services database (WFIGS)\footnote{\url{https://data-nifc.opendata.arcgis.com/datasets/nifc::wfigs-current-interagency-fire-perimeters/about}, accessed on 1 February 2026}.
 The data contains polygons representing the mapped perimeters of wildfire events across the U.S. 
 The locations of the fire events are shown in Figure ~\ref{fig:fire_perimeter_map}. Further details regarding the 51 fire events appear in 
 Table \ref{tab:fire_events} in Appendix~\ref{app:Historical Fire Events}. 
 
\section{Methodology}
\label{sec:methodology}
As detailed below, the construction of the ML fire detection algorithm consists of the following steps:
\begin{enumerate}[label=(\roman*)]
    \item Spatial assignment of VIIRS fire detections to the GOES ABI grid (Section \ref{subsec:Rasterizing VIIRS}).
    \item Construction of the labeled dataset (Section \ref{subsec:Training Dataset Construction}).
    \item ML model training (Section \ref{subsec:Model training}).
    \item ML model evaluation (Section \ref{subsec:Model evaluation}).
\end{enumerate}
Code of the trained ML model and some example data are available at the author's GitHub repository\footnote{\url{https://github.com/asafvanunu/GOES_wildfire_detection}, accessed on 1 February 2026}.

\subsection{Assigning VIIRS fire detections to GOES ABI grid}
\label{subsec:Rasterizing VIIRS}
\begin{figure}[t]
    \centering
    \includegraphics[width=0.85\linewidth]{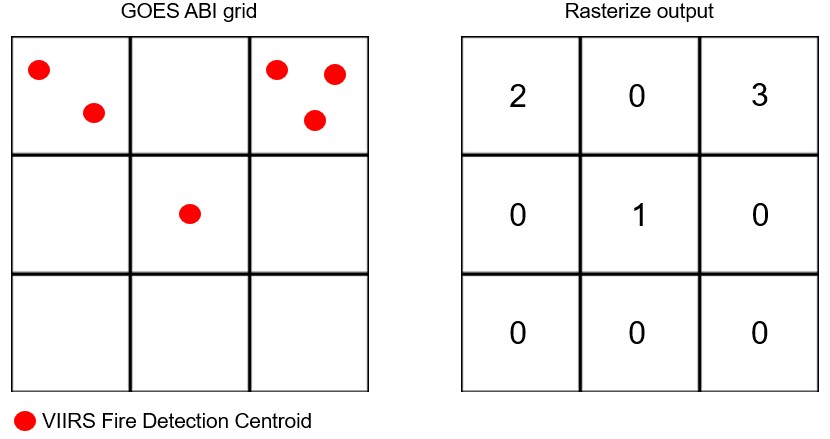}
    \caption{Rasterizing VIIRS fire detections onto the GOES ABI grid.}
    \label{fig:rasterize}
\end{figure}

As mentioned above, VIIRS and GOES ABI have different spatial resolutions and grid systems.
Thus, to label GOES ABI pixels as fire/no fire, VIIRS fire detection pixel centroids were rasterized onto the GOES ABI fixed grid. The rasterized output is a GOES ABI grid, in which each pixel represents the number of VIIRS fire-detection centroids that intersect it. Pixels without any VIIRS fire-detection intersections were assigned a value of 0. Figure ~\ref{fig:rasterize} illustrates the rasterize function, available in the \texttt{Rasterio} Python library\footnote{\url{https://rasterio.readthedocs.io/en/stable/api/rasterio.features.html}, accessed on 1 February 2026} \citep{gillies2019rasterio}.
%\paragraph{VIIRS Bow-tie Pixels Removal}\mbox{}\\

The VIIRS instrument captures imagery in successive swaths known as scan lines.
Due to its technical properties, there is an overlap between VIIRS pixels from adjacent scan lines.
This phenomenon is commonly referred to as the bow-tie effect \citep{seaman2015visible}.
Most of these duplicated pixels are removed onboard the instrument.
However, residual bow-tie pixels remain and may affect the analysis.
Thus, duplicated VIIRS pixels were removed using the method described by \citet{Vanunu2025_FireMisalignments}.     

\subsection{Labeled dataset construction}
\label{subsec:Training Dataset Construction}
Next, we describe the process to label GOES ABI pixels as fire/no fire. These steps include handling spatial misalignments between VIIRS and GOES ABI fire detections, labeling non-fire pixels, and extracting input features for the ML model.
%\paragraph{Misalignment aware labeling}\mbox{}\\
A significant challenge in constructing a labeled dataset is the frequent spatial misalignment between VIIRS fire detections and GOES ABI observations.
As shown in several studies \citep{PESTANA_parallax_2022, Ayala_Parallax_2023, Vanunu2025_FireMisalignments}, there can be displacements of several kilometers between GOES ABI and VIIRS fire pixel locations.
This, in turn, can significantly affect the accuracy of the trained ML model.

\begin{figure}[t]
    \centering
    \includegraphics[width=0.85\linewidth]{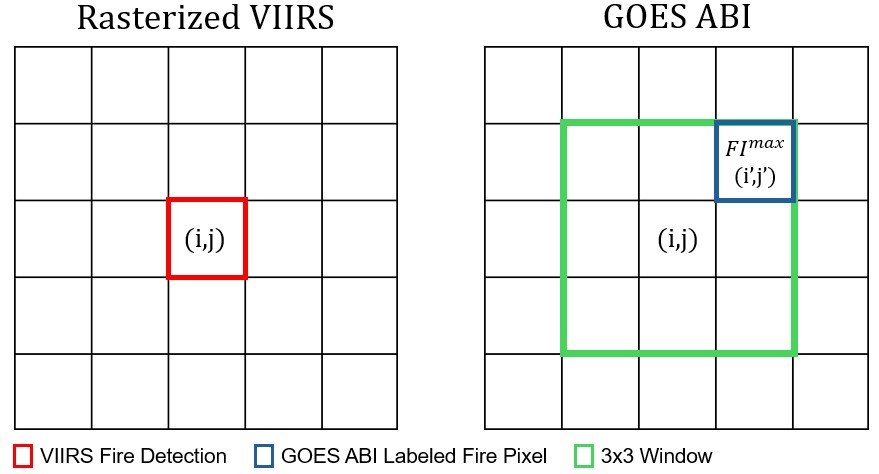}
    \caption{Illustration of the fire labeling process. Left: VIIRS fire detection at pixel $(i,j)$ in the rasterized VIIRS matrix $V$. Right: A $3 \times 3$ window is applied in the GOES MCMI matrix $G$. The pixel $G(i',j')$ with the highest FI value is labeled as fire.}
    \label{fig:labeling_window}
\end{figure}

To ensure accurate labeling of GOES ABI fire pixels that takes into account spatial misalignments, a buffer window was applied.
Specifically, let $V$ be the rasterized image. A value $V(i,j)\geq1$ represents a VIIRS fire detection at ABI grid location $(i,j)$ in the rasterized output (Section~\ref{subsec:Rasterizing VIIRS}).
Similarly, let $G(i,j)$ denote a pixel at the same location and time stamp in the GOES MCMI image.
As illustrated in Figure \ref{fig:labeling_window}, for each location $(i,j)$ with $V(i,j)\geq1$, a $3 \times 3$ buffer window was applied around $G(i,j)$.
Within this window, the GOES ABI pixel with the highest fire signal, quantified by the Fire Index (FI), was labeled as a fire pixel.
The FI \citep{Zhao_2022_DL_GOES} is the normalized ratio between the shortwave infrared (SWIR) band at 3.9~µm and the thermal infrared (TIR) band at 11.2~µm:
\begin{equation}
FI = \frac{B_{3.9} - B_{11.2}}{B_{3.9} + B_{11.2}}
\label{eq:fire_index}
\end{equation}
where $B_{3.9}$ and $B_{11.2}$ are the GOES ABI brightness temperatures (in Kelvin) at 3.9~µm and 11.2~µm bands, respectively. The 3.9~µm band is sensitive to fires, while the 11.2~µm band is not. High FI values are associated with active fires and thus considered a reliable indicator of fire presence\citep{Zhao_2022_DL_GOES}. 

Note that applying a (square or circular) buffer window around a GEO fire detection is common practice in evaluating the accuracy of GEO fire products \citep[e.g.,][]{Freeborn_validation, Wickramasinghe_validation, XU_buffer_2017, XU_Buffer2021, HALL_2019_GOES_SEVIRI, Hall_2023_GOES_H8}.
If a LEO fire detection falls within this window, the GEO fire detection is considered a valid fire. Here, in contrast, the buffer window is placed around the VIIRS fire detection, and the purpose is different: to label GEO pixels as fires while accounting for and correcting potential misalignments.

%\paragraph{Non-fire pixel labeling}\mbox{}\\
For non-fire pixel labeling, from each GOES ABI image, 500 pixels that satisfy the following two requirements were selected: 
(1) It cannot intersect a labeled fire pixel in a $3 \times 3$ window.
(2) It cannot be within a $3\times3$ window of any pixel with $V(i,j)>1$ in the corresponding image $V$.
These requirements ensure non-fire pixels are distant from fire-labeled pixels.

%\paragraph{Input features for the ML algorithm}\mbox{}\\
For each labeled pixel (fire or non-fire), the values of all its 16 bands as well as its FI value were extracted.
Additionally, a $3 \times 3$ window was applied around each pixel to obtain its spatial context.
From the pixels within the window (excluding the center pixel), five statistics were computed for all 17 bands (including the FI): mean, median, minimum, maximum, and standard deviation.
Pixels flagged as cloud-covered by the ACM product were excluded from these calculations. If all of the neighboring pixels were flagged as clouds, a fill value of -999 was assigned.
The number of contextual features is $5 \times 17 = 85$.
In total, there are 102 features (17 band values at the center pixel and 85 contextual features).

%\paragraph{Data filtering}\mbox{}\\
To exclude small fires that the GOES ABI struggles to detect, two filtering criteria were applied to the fire-labeled pixels:
(1) The number of VIIRS fire detections in the $3\times3$ window around the center pixel at the corresponding matrix $V$ was required to be $\geq 2$; and
(2) The FI value was required to be $> 0$. 

Next, the class imbalance between fire and non-fire pixels was addressed.
Satellite-based wildfire detection datasets are highly imbalanced, as non-fire pixels greatly outnumber fire pixels.
To mitigate this, a 1:20 ratio of fire to non-fire pixels was applied.
This ratio was chosen to preserve sufficient background pixels while preventing the non-fire class from dominating the model training.
Ratios of 1:10 and 1:30 were also tested. Although the 1:30 ratio achieved similar precision, the 1:20 ratio was selected because its better F1 score.
In summary, the training dataset contained 202,000 fire-labeled and 2,165,000 non-fire-labeled GOES ABI pixels. After the filtering criteria were applied, 65,877 fire-labeled and 1,317,540 non-fire-labeled pixels remained.

\subsection{Model training}
\label{subsec:Model training}
In this study, a CatBoost~\citep{catboost} algorithm was trained to detect fires. CatBoost (Categorical Boosting) is a gradient-boosting method that builds an ensemble of decision trees. Each tree is trained to correct the errors made by the earlier trees.

CatBoost can process different types of input features without complex pre-processing. It handles numerical values and categorical variables in a unified way. Unlike traditional gradient boosting 
frameworks, CatBoost incorporates specific algorithmic innovations to improve accuracy and generalization. These include: (1) ordered boosting, a technique that shuffles training data to prevent the model from memorizing the target ('target leakage'); (2) native handling of categorical variables, which avoids the need for complex pre-processing of non-numerical data; and (3) oblivious decision trees, a balanced tree structure that acts as a regularizer to prevent overfitting and significantly speeds up prediction times. 
CatBoost has been widely applied across a variety of domains~\citep{hancock2020catboost}, 
including remote sensing~\citep{boosting_in_RS}.
The training dataset was randomly divided into 80\% for training and 20\% for validation. Further details regarding the CatBoost model training are provided in 
Appendix ~\ref{app:CatBoost Model Appendix}.

\subsection{Accuracy evaluation of a fire detection algorithm}
\label{subsec:Model evaluation}
\begin{figure}[t]
    \centering
    \includegraphics[width=0.85\linewidth]{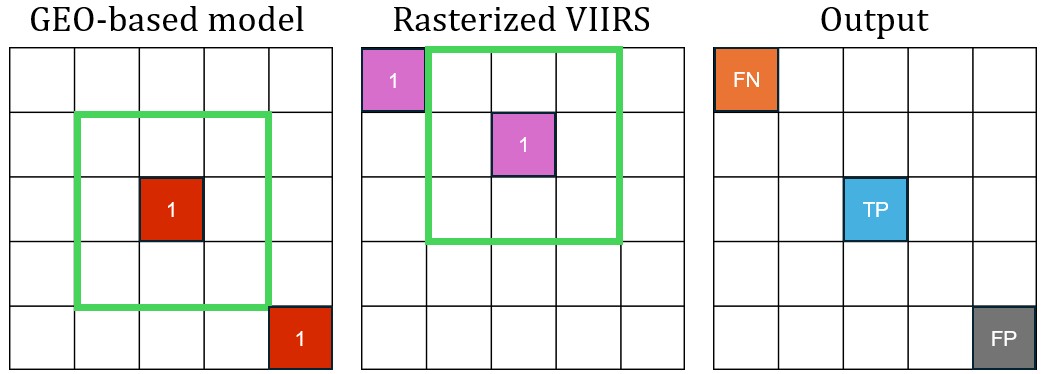}
    \caption{Illustration of accuracy evaluation of a GEO-based fire detection algorithm using a $3 \times 3$ window. (left) Red pixels are GEO-based fire detections; (center) Pink pixels are rasterized VIIRS fire detections; (right) FN, TP, and FP are false negative, true positive, and false positive, respectively.}
    \label{fig:evaluation_window}
\end{figure}

%\paragraph{Misalignment aware evaluation}\mbox{}\\
To evaluate the accuracy of a fire detection algorithm, spatial misalignments between GOES ABI and VIIRS observations were considered. 
Similar to Section \ref{subsec:Training Dataset Construction}, let $M(i,j)=1$ denote that at ABI pixel location $(i,j)$ the fire detection algorithm predicted a fire.  
A pixel $(i,j)$ with $M(i,j)=1$ is classified as a True Positive (TP) if there is at least one $V(i',j')\geq1$ within a $3 \times 3$ window around $(i,j)$.
Otherwise, the pixel is considered a False Positive (FP). 
A False Negative (FN) pixel is defined as follows.
If the $3 \times 3$ window around $V(i,j)\geq1$ does not contain a model fire detection, then the pixel is defined as FN.
The calculation of FN, TP, and FP with a $3 \times 3$ window is illustrated in Figure \ref{fig:evaluation_window}. The red pixels are the model fire detections. In contrast, the pink pixels are rasterized VIIRS fire detections.

%\paragraph{Accuracy metrics}\mbox{}\\
The accuracy of a fire detection algorithm is evaluated by its recall, precision, and F1 scores.
These are standard metrics for classifier assessment.
In the context of fire detection, recall is the proportion of true fire pixels that the algorithm correctly identifies as fires.
Precision is the ratio of correctly detected fire pixels to the total number of flagged fire pixels by the algorithm.
The harmonic mean of precision and recall is the F1 score. All three metrics range from 0 to 1, where higher values indicate improved accuracy.

To compute recall, precision, and F1, the total number of TP, FP, and FN pixels was summed across the entire image dataset.
We denote these quantities by $\text{TP}^\text{total}$, $\text{FP}^\text{total}$ and $\text{FN}^\text{total}$. Accordingly, let $n$ denote a pair of GEO/rasterized VIIRS images, and let $i$ represent the index of a single image. For each image $i$, we denote by $TP_i$, $FP_i$, and $FN_i$ the number of pixels in that image classified as TP, FP, and FN, respectively.
The totals were then calculated:   
\begin{equation*}
    \text{TP}^{\text{total}} = \sum_{i=1}^{n} \text{TP}_i \qquad \text{FP}^{\text{total}} = \sum_{i=1}^{n} \text{FP}_i \qquad \text{FN}^{\text{total}} = \sum_{i=1}^{n} \text{FN}_i
\end{equation*}
Based on these totals, the precision, recall, and F1 metrics were calculated:
\begin{align}
    \text{Recall} &= \frac{\text{TP}^{\text{total}}}{\text{TP}^{\text{total}} + \text{FN}^{\text{total}}} \\
    \text{Precision} &= \frac{\text{TP}^{\text{total}}}{\text{TP}^{\text{total}} + \text{FP}^{\text{total}}} \\
    \text{F1} &= \frac{2 \times \text{Precision} \times \text{Recall}}{\text{Precision} + \text{Recall}}
\end{align}

\section{Results}
This section presents the empirical evaluation results of the CatBoost model trained as described in section \ref{subsec:Model training}.
In particular, we compare its performance with that of the GOES FDC in terms of accuracy, early-detection capabilities and ability to detect small fires.

\subsection{CatBoost performance against the GOES FDC}
\label{subsec: Model Performance}
\begin{figure}[t]
    \centering
    \includegraphics[width=0.6\textwidth]{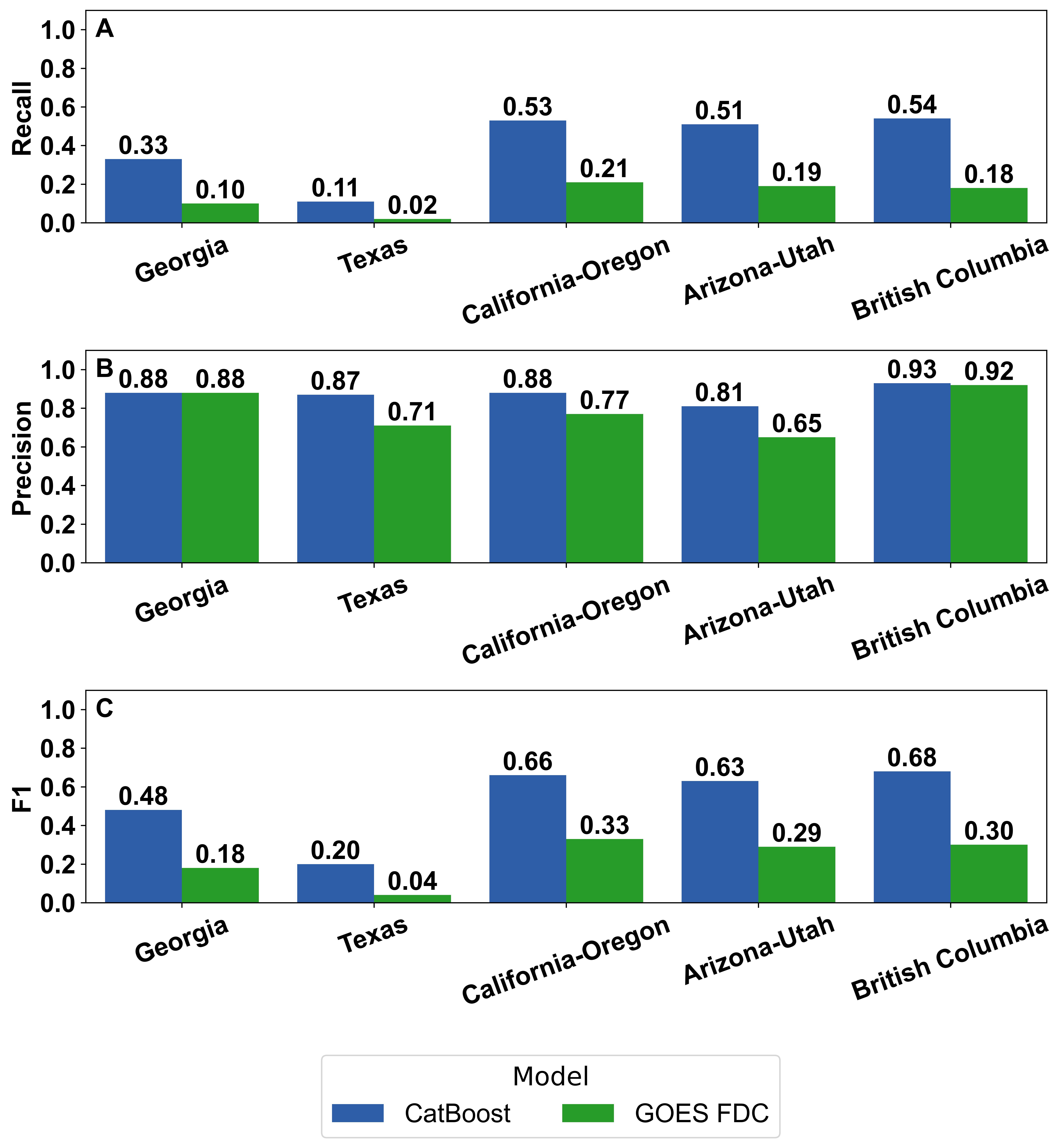}
    \caption{Comparison of performance metrics in terms of recall, precision, and F1 between the CatBoost model and the GOES FDC product in five different evaluation regions.
    }
    \label{fig:evaluation}
\end{figure}

Figure~\ref{fig:evaluation}a-c shows the recall, precision, and F1 scores, respectively, for the CatBoost model and the GOES FDC product across all five evaluation regions (Georgia, Texas, California-Oregon, Arizona-Utah and British Columbia).
The results show that the CatBoost model outperformed the GOES FDC product across all regions and evaluation metrics. It achieved lower false alarm rates and equal or higher detection rates.
In California–Oregon, Arizona-Utah and British Columbia, CatBoost performed better in recall, precision, and F1, regions that are not in the training dataset.
This demonstrates the transferability of the trained CatBoost model to regions outside the training areas.
In all regions, the CatBoost ML model achieved F1 scores that were 0.16-0.38 higher than those of the GOES FDC.

\begin{figure}[t]
    \centering
    \includegraphics[width=0.8\textwidth]{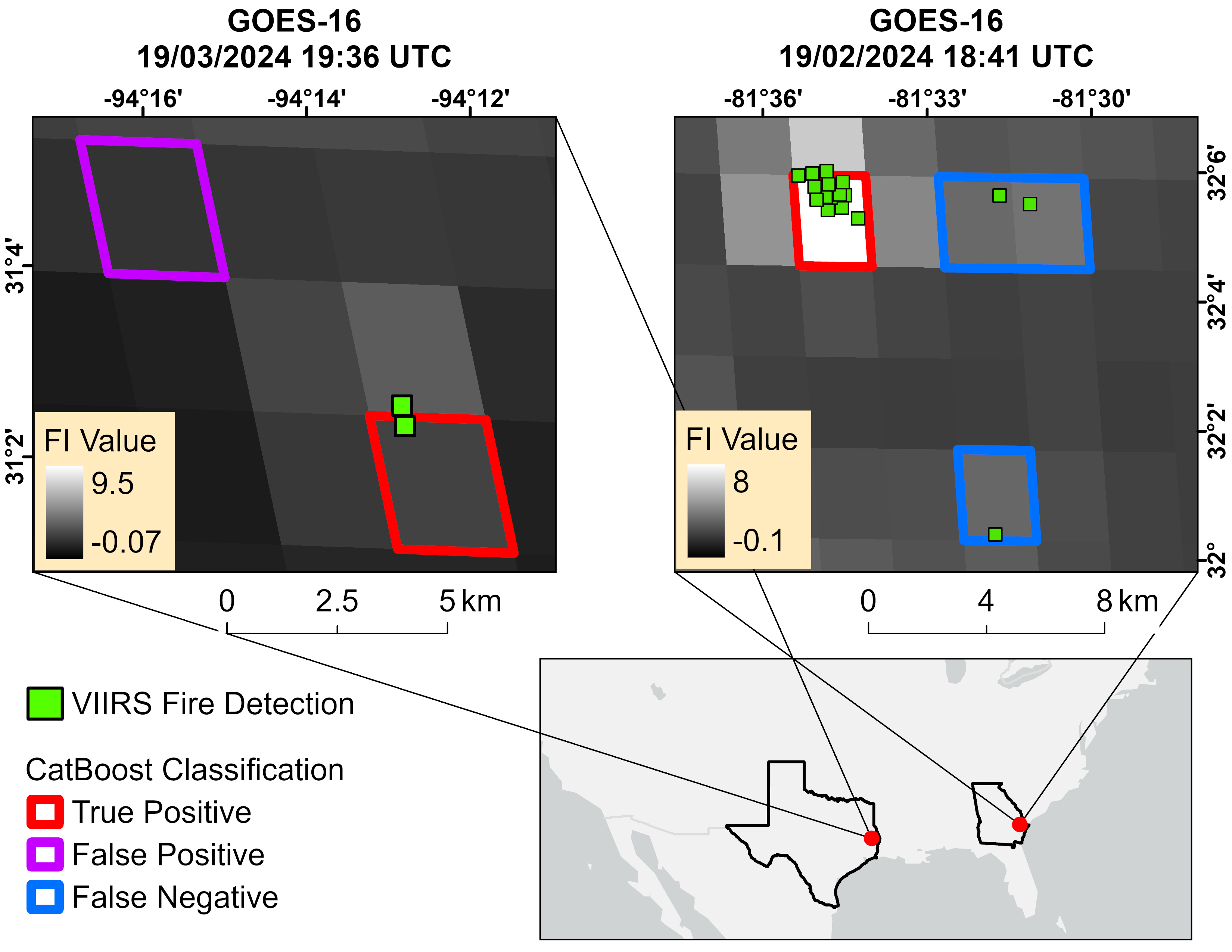}
    \caption{Example of CatBoost model predictions on the background of GOES-16 ABI FI images in Texas (left) and Georgia (right).
    VIIRS 375 m fire detections are shown as green squares.
CatBoost prediction pixels are color-coded in red, purple, and blue for TP, FP, and FN, respectively.}
    \label{fig:FN_FP_map}
\end{figure}

%\paragraph{Visual example}\mbox{}\\
Going beyond the quantitative evaluation described above, it is instructive to look at two specific fire events, one in Texas and the other in Georgia. Figure ~\ref{fig:FN_FP_map} shows the CatBoost model's predictions.
The green squares of width 375 m, are fires detected by the VIIRS fire product on images of the two regions taken at matching times (up to $\pm~2.5$ min).
The left image is of Texas on 19 March 2024 at 19:36 UTC, and the right image is of Georgia on 19 February 2024 at 18:41 UTC. CatBoost pixels are marked in red for TP, purple for FP, and blue for FN. In Texas, the FP pixel is near the TP, suggesting that proximity to the heat source may have caused misclassification. In Georgia, small fires with single VIIRS detections were missed by the model and labeled as FN.

\subsection{Daytime/nighttime performance}
Following the evaluation of the CatBoost model against the GOES FDC, it is important to explore how the time of day impacts detection capabilities. In this section, we compare the accuracy fire detection accuracy of the CatBoost and FDC models during the day and during the night across all study regions. The separation between day and night images was determined according to the sunrise and sunset times for each region. These times were computed based on the longitude, latitude, and date of each image using the Sunrise-Sunset API\footnote{https://sunrise-sunset.org/api, accessed on 1 February 2026}. Table \ref{tab:day_night_images} details the separation of the full evaluation dataset into daytime and nighttime images. The dataset contains a relatively balanced number of day and night images for most regions.

\begin{table}[t]
\centering
\caption{Number of daytime and nighttime GOES ABI images for all study regions.}
\vspace{0.5em}
\label{tab:day_night_images}
\begin{tabular}{l c c}
\hline
\textbf{Region} & \textbf{Daytime images} & \textbf{Nighttime images} \\
\hline
Georgia           & 101 & 84  \\
Texas             & 153 & 172 \\
California-Oregon & 180 & 183 \\
Arizona-Utah      & 135 & 150 \\
British Columbia  & 38  & 77  \\
\hline
\end{tabular}
\end{table}

\begin{figure}[H]
    \centering
    \includegraphics[width=\textwidth]{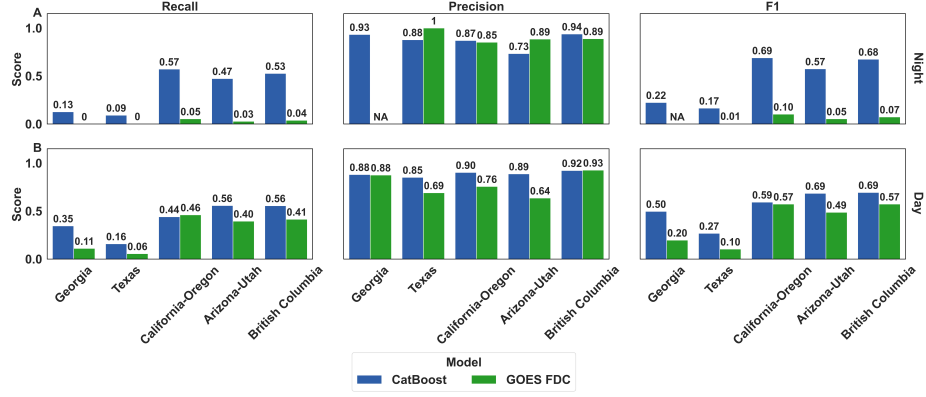}
    \caption{Recall, precision, and F1 scores of the CatBoost model and GOES FDC across all study regions. Panel A (upper) nighttime and Panel B (lower) daytime.
    }
    \label{fig:day_night_eval}
\end{figure}

Figure ~\ref{fig:day_night_eval} shows the recall, precision, and F1 scores of the CatBoost and GOES FDC in all study regions during nighttime (Figure ~\ref{fig:day_night_eval}a) and during daytime (Figure ~\ref{fig:day_night_eval}b). During the daytime, recall and precision are higher for the CatBoost model, except for recall in California-Oregon and precision in Georgia and British Columbia, which are almost identical between the two algorithms. For the F1 score, CatBoost is higher than the FDC in all study regions during the day.

During the nighttime, in contrast, the CatBoost algorithm is significantly more accurate than the FDC. Notably, the recall score of CatBoost is significantly higher than the FDC. In fact, the FDC recall is 0 in Georgia and Texas, and it is almost 0 in the other regions. This means the FDC fails to detect almost all fires in the nighttime images of the study dataset. In Georgia, precision could not be calculated because the FDC did not detect any true or false positives across all 84 images. In Texas and Arizona-Utah, the FDC precision was higher than CatBoost because of the low number of FDC detections. There were only 13 fire detections in 172 images in Texas, while there were 4,409 GOES ABI pixels with at least one VIIRS fire detection in them. Similarly, there were only 26 fire detections in 134 images in Arizona-Utah, compared to 850 GOES ABI pixels with at least one VIIRS fire detection. For comparison, the CatBoost model in the corresponding regions had 477 and 691 fire detections, respectively. In the remaining regions, CatBoost achieves higher precision. For the nighttime, the CatBoost F1 scores were significantly higher, ranging from 0.17 to 0.69, while the FDC ranged from only 0.01 to 0.1. Overall, these results demonstrate that the GOES FDC struggles to detect fires at night, and the CatBoost model provides superior nighttime detection capabilities.

\subsection{Impact of training dataset size on CatBoost performance}
\label{subsec: Model Performance catboost models}

To demonstrate the importance of a large and diverse labeled dataset, we compared the model described in section \ref{subsec:Model training} with the following two additional models. The first is a CatBoost model trained on a random subset of only 2,000 fire pixels and 20,000 non-fire pixels. The second model was trained on 2,000 fire pixels and 20,000 non-fire pixels drawn only from April and May.

\begin{figure}[!ht]
    \centering
    \includegraphics[width=\textwidth]{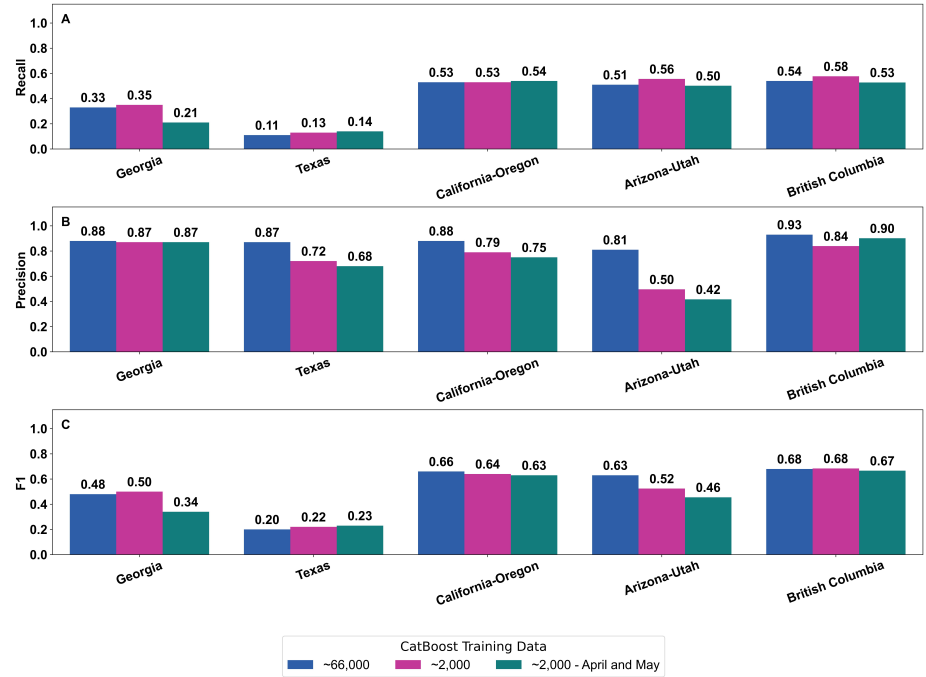}
    \caption{Comparison of performance metrics between CatBoost models using different number of fire pixels for training in five different evaluation regions.
    }
    \label{fig:evaluation_only_cat}
\end{figure}

Figure \ref{fig:evaluation_only_cat}a-c presents precision, recall, and F1 scores, respectively, for all five models across Georgia, Texas, California-Oregon, Arizona-Utah and British Columbia. The full model, trained on 66,000 fire pixel, shows substantially higher precision in Texas, California–Oregon and Arizona-Utah than the two reduced-size models, while all models display similar precision in Georgia. Precision in Texas is 0.87 for the 66,000-pixel model, which is significantly higher than 0.72 and 0.68 achieved by the two 2,000-pixel models. In California-Oregon, precision is 0.88 for the 66,000-pixel model and 0.79 and 0.75 for the smaller models. In Arizona-Utah, the precision is 0.81 for the 66,000-pixel model, compared to 0.5 and 0.42 for the small models. These results show that small training datasets lead to higher false-alarm rates, especially when the dataset is restricted to only some months of the year.

Recall values are generally similar across models, with differences of about 0.04. The main exception is Georgia, where the April–May model produces a recall of only 0.21, compared with 0.33 and 0.35 for the 66,000-pixel and 2,000-pixel models. This result highlights the potential limitations of using a dataset that covers only a few specific months.

F1-scores are also similar across most models and regions, apart from the April–May model in Georgia and Arizona-Utah, which shows a clear drop in performance. In California–Oregon and Arizona-Utah, which were not part of the training area, the highest F1-score is achieved by the 66,000-pixel model. This indicates better transferability and generalization when the training dataset is large and diverse.

In summary, the 66,000-pixel model produces far fewer false-alarms than the reduced training size models while maintaining similar recall. Training on a small or limited dataset can lead to increased false-alarms and decreased performance in regions outside the training domain.

\subsection{Early detection capability}
\label{subsec:First time detection}
Early fire detection is crucial for extinguishing small fires before they escalate to large and destructive wildfires.
As mentioned in Section \ref{subsec:Data}, a dataset of 51 fires with their fire perimeters was constructed to compare the early detection capabilities of the CatBoost model and the GOES FDC. 
First, for each fire event in the dataset, CatBoost model predictions and the corresponding FDC product were collected for all GOES ABI images acquired within the 8 hours preceding and 2 hours following the first VIIRS fire detection. Next, the images were cropped to a 10~km buffer around the fire polygon.
Lastly, the time difference between the first fire detection of the CatBoost model and the GOES FDC was computed (in minutes).

\begin{figure}[H]
    \centering
    \includegraphics[width=\textwidth]{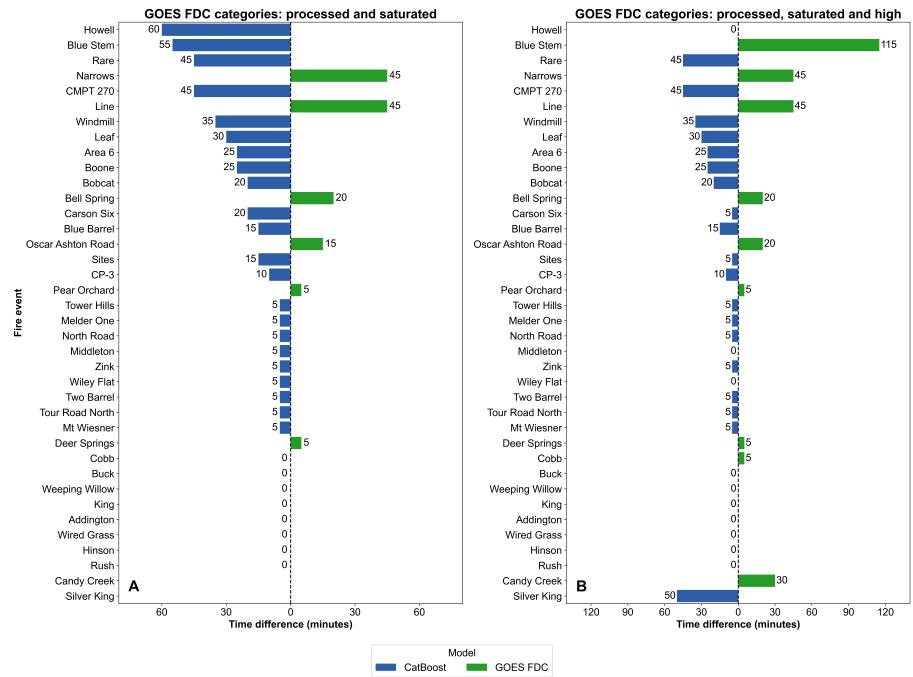}
    \caption{First detection time difference between CatBoost and GOES FDC across fire events. Panel A shows the analysis using only the \texttt{processed} and \texttt{saturated} FDC categories, and Panel B includes the \texttt{high} category. Fire events that were detected simultaneously by CatBoost and GOES FDC have a time delta label of "0".}
    \label{fig:first_time_detection}
\end{figure}

Figure ~\ref{fig:first_time_detection} summarizes the time difference between the CatBoost and GOES FDC for 38 fire events where both methods detected the fire before the time of the VIIRS image first detection. Figure ~\ref{fig:first_time_detection}a (left) shows the results using only the \texttt{processed} and \texttt{saturated} FDC categories, while Figure ~\ref{fig:first_time_detection}b (right) includes also the \texttt{high} confidence category. The x-axis shows the time difference in minutes, and the y-axis lists the fire events. In ~\ref{fig:first_time_detection}a, 22 fire events were first detected by CatBoost, with a mean earlier detection time of 20 min and a standard deviation of 17 min, while GOES FDC first detected only six fires. In ~\ref{fig:first_time_detection}b, the FDC fire detection capability improved with the addition of the \texttt{high} category addition. The CatBoost first detected 19 fires with a mean earlier detection time of 18 min and a standard deviation of 16 min. Here the GOES FDC first detected nine fires. For both analyses, 10 fire events were detected simultaneously by the CatBoost model and GOES FDC and were therefore labeled as "0".

A deeper examination of these results reveals several notable cases, specifically the Howell, Blue Stem, Candy Creek, and Silver King fires. For the Howell fire, CatBoost detected the event 60 minutes earlier than the GOES FDC when using only the \texttt{processed} and \texttt{saturated} categories. However, when the \texttt{high} category was included, the time difference became zero. In the Blue Stem fire, CatBoost initially detected the fire 55 minutes before the FDC. Yet, with the addition of the \texttt{high} category, the FDC detected the fire 115 minutes earlier than CatBoost. This indicates that the \texttt{high} category can sometimes enable the FDC to detect fires simultaneously with or earlier than the CatBoost model. Furthermore, the Candy Creek and Silver King fires were not detected at all by the FDC when restricted to the \texttt{processed} and \texttt{saturated} categories. Once the \texttt{high} category was added, both fires were detected by the FDC, and for the Candy Creek fire, the FDC even detected it 30 minutes earlier than CatBoost. These examples demonstrate that FDC categories with higher false alarm rates can, in certain situations, provide better early fire detection capabilities than the \texttt{precessed} and \texttt{saturated} categories.

Table~\ref{tab:first_time_detection_table} describes the remaining 13 fire events split into two categories: (1) The CatBoost model detected the fire and GOES FDC did not detect it before VIIRS; and (2) Neither CatBoost nor FDC detected a fire before the time of the VIIRS image. 

Overall, out of 51 fire events, CatBoost detected 26 fires before both VIIRS and GOES FDC in the first analysis, and 23 in the second analysis. Ten fires were detected simultaneously by CatBoost and GOES FDC, while nine were first detected by VIIRS. GOES FDC first detected six fires in the first analysis and nine fires in the second. These results demonstrate that the CatBoost model has substantially better early detection capabilities than the GOES FDC product.

\begin{table}[H]
\centering
\caption{Summary of fire events detected (1) Only by CatBoost and VIIRS; (2) Only by VIIRS.}
\vspace{0.5em}
\label{tab:first_time_detection_table}
\begin{tabular}{|l|c|c|}
\hline
\textbf{Fire event} & \textbf{First fire detection} & \textbf{Time difference (minutes)} \\
\hline
Beech            & \multirow{4}{*}{CatBoost} & 300 \\
Bentley          &                           & 480 \\
Davis            &                           & 445 \\
Smokehouse Creek &                           & 245 \\
\hline
Airport          & \multirow{9}{*}{VIIRS}    & \multirow{9}{*}{No CatBoost detection} \\
Blue Mountain    &                           &  \\
Boise            &                           &  \\
Dry Branch       &                           &  \\
Duck Nest        &                           &  \\
Fall Branch      &                           &  \\
Fodderstack      &                           &  \\
Foote            &                           &  \\
Walker           &                           &  \\
\hline
\end{tabular}
\end{table}

\subsection{Fire detection probability}
\label{subsec:Fire Detection Probability}
This section analyzes the CatBoost and GOES FDC fire detection probability as a function of the number of VIIRS fire detections within a spatial window of size \(3 \times 3\) around the GEO pixel predicted as fire by the corresponding method. 
This was done to assess the models' sensitivity to different fire sizes. The approach presented here is similar to those conducted by \citet{Vanunu2025_FireMisalignments}.

\begin{figure}[H]
    \centering
    \includegraphics[width=0.6\textwidth]{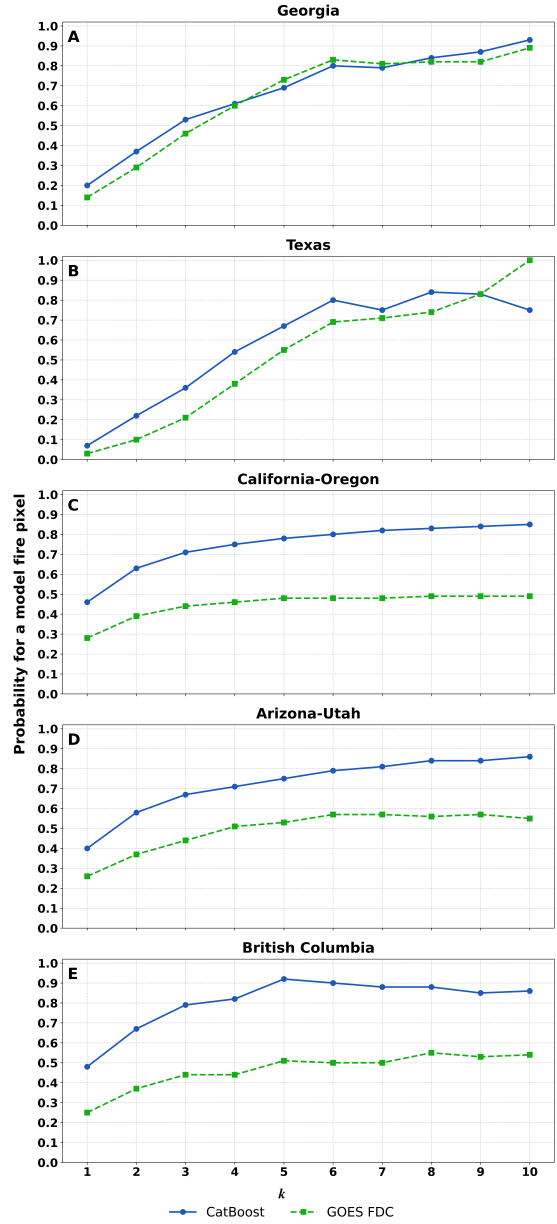}
    \caption{Probability of a fire pixel detected by the CatBoost and GOES FDC as a function of the minimum number of VIIRS fire detections inside a GOES ABI pixel (\(k\)). Evaluation regions: (A) Georgia, (B) Texas, (C) California-Oregon, (D) Arizona-Utah and (E) British Columbia.}
    \label{fig:fire_detection_probability}
\end{figure}

Let \( M \) denote the binary image representing the output of a fire detection algorithm.
A pixel value of 1 indicates a detected fire, and 0 otherwise.
As described in Equation \ref{eq:neighborhood}, for the window parameter \( w\) ($w=1 \text{ for a } 3\text{x}3\text{ window}$), the neighborhood set around pixel \((i,j)\) is defined as:
\begin{equation} \label{eq:neighborhood}
N_{ij} = \{ (i', j') \mid \max(|i' - i|, |j' - j|) \leq w \}
\end{equation}
For different values of an integer \( k \), the total number of pixels \( N_{TP,k} \) with at least \( k \) VIIRS fire detections and a model fire detection within a distance \( w \) was calculated. This calculation was performed for all images in the dataset.
Accordingly, for each image \(M\), let \( V \) represent the corresponding rasterized VIIRS output with threshold \( k \)
\begin{equation} \label{eq:truepos}
N_{TP,k} = \sum_{\text{images}} \sum_{(i,j)} 1 \left[ V(i,j) \geq k \text{ and for some } (i', j') \in N_{ij}, \; M(i', j') = 1 \right]
\end{equation}
Likewise, the number of FP pixels \( N_{FN,k} \) was calculated
\begin{equation} \label{eq:falseneg}
N_{FN,k} = \sum_{\text{images}} \sum_{(i,j)} 1 \left[ V(i,j) \geq k \text{ and for all } (i', j') \in N_{ij}, \; M(i', j') = 0 \right]
\end{equation}
Hence, \( N_{TP,k} + N_{FN,k} \) equals the total number of pixels containing at least \( k \) VIIRS fire detections. The probability that in a $w$ window around each such pixel, there is a model fire detection was calculated.
\begin{equation} \label{eq:probability}
P \left( M(i', j') = 1 \text{ for at least one } (i', j') \in N_{ij} \mid V(i,j) \geq k \right) = \frac{N_{TP,k}}{N_{TP,k} + N_{FN,k}}
\end{equation}
Figure ~\ref{fig:fire_detection_probability} shows the estimated probabilities as a function of
$k$ across the five evaluation regions: Georgia (Figure ~\ref{fig:fire_detection_probability}a), Texas (Figure ~\ref{fig:fire_detection_probability}b), California-Oregon (Figure ~\ref{fig:fire_detection_probability}c), Arizona-Utah (Figure ~\ref{fig:fire_detection_probability}d), and British Columbia (Figure ~\ref{fig:fire_detection_probability}e). 
The x-axis \(k\) represents the minimum number of VIIRS fire detections contained within a single GOES ABI pixel.

In California-Oregon, Arizona-Utah and British Columbia, the CatBoost model yields substantially higher fire-detection probabilities than those of GOES FDC. For instance, in California-Oregon at \(k=1\), GOES FDC has a fire probability of 0.28 while CatBoost’s is 0.46. At \(k=5\), the CatBoost probability is 0.78 compared to only 0.48 for GOES FDC.
In Texas, the CatBoost model shows higher probabilities than GOES FDC for thresholds below \(k=10\).
At \(k=10\), the GOES FDC probability exceeds that of the CatBoost model. This can be attributed to the very small sample size of only eight ABI pixels that contained ten or more VIIRS fire detections.
Overall, the CatBoost model consistently achieves higher probabilities than GOES FDC across all regions, particularly in California and Oregon.
Hence, in most cases, the CatBoost model is more sensitive to small fires and is more likely to detect them.

\section{Discussion and conclusions}
In this study, a CatBoost model was trained for wildfire detection using a large, carefully annotated dataset of GOES-16 ABI imagery. CatBoost was selected for this task because it is a modern, highly efficient gradient boosting framework that is relatively simple to train.
The model was evaluated and compared to the GOES FDC product. For both training and evaluation, VIIRS fire detections served as a reference.
The analysis in this study focused on prediction accuracy, early-detection capability, and fire-detection probability.

The results demonstrate that the CatBoost model consistently outperforms GOES FDC across all five regions considered in the study. It achieved higher precision, recall, and F1-scores. Recall values were relatively low for both models. This is due to the coarse spatial resolution of ABI compared to VIIRS. As a result, small fires detected by VIIRS were classified as false negatives by both the GOES FDC as well as by the CatBoost model. Despite this limitation, CatBoost achieved a higher recall than GOES FDC. The superior performance of the CatBoost model in California–Oregon, Arizona-Utah and British Columbia indicates generalization ability to regions outside the training domain. These evaluation regions feature different climatic conditions and varying satellite view angles compared to the training data. Additional daytime/nighttime evaluation between the CatBoost and the GOES FDC revealed striking differences. While CatBoost maintained reliable nighttime detection, the GOES FDC nighttime recall dropped to near zero across all regions. This highlights the CatBoost capabilities for fire detection across all day hours.
In a different analysis, CatBoost models trained on smaller datasets highlighted the importance of training on a large, diverse labeled dataset for reliable detection and robust performance in regions outside the training area. Models trained on only 2,000 fire pixels exhibited substantially higher false-alarm rates and weaker transferability.
These findings have important implications for operational wildfire monitoring. Accurate detection is critical for rapid response, especially in remote regions. In addition, improved GEO-based detection can bridge the temporal gap between LEO satellite overpasses, providing continuous fire monitoring.

The results are consistent with previous studies on ML-based fire detection using GEO satellites. \citet{KANG_2023_Toward_H8} trained a GEO-based deep learning fire detection model using imagery from the Japanese Himawari-8 Advanced Himawari Imager (AHI). The model was trained over regions in Asia and demonstrated improved performance compared to the operational Himawari fire product. The reported F1 scores were 0.66 and 0.50 at in situ validation sites in Asia.
In addition, transferability analysis over Australia yielded an average F1 score of 0.68. 
These F1 score ranges are comparable to those achieved in the current study. 
Overall, their results showed promising potential for GEO-based ML fire-detection models applied across large geographic regions, potentially outperforming existing GEO fire products.

Despite these advancements in the Asia–Pacific region, the application of machine-learning-based fire-detection methods to the GOES-ABI platform remains limited. To date, only one study has applied machine learning to GOES-ABI-based fire detection. \citet{Zhao_2022_DL_GOES} trained a deep learning model on GOES-ABI imagery using a relatively small dataset constructed from only four major fire events. Reported F1 scores were 0.71, 0.63, and 0.18 for the Creek, Camp, and Doctor Creek fires, respectively. Their model detected 24 of 38 fires earlier than VIIRS, comparable to the early-detection capability observed in the present results. However, their evaluation was limited to a small number of individual fire events and did not include a comparison with the operational GOES Fire Detection and Characterization product.

While \citet{Zhao_2022_DL_GOES} demonstrated the potential of deep learning for GOES ABI fire detection, our work extends their study in several important ways. We evaluated the model across large-scale geographic regions rather than focusing on a small number of specific large fire events. In addition, we benchmark our model against the GOES FDC, and to the best of our knowledge, this study is the first to do so. We also explicitly assessed early detection relative to both VIIRS and GOES FDC. In addition, we quantified the detection probability as a function of VIIRS fire detections, an aspect that has not been extensively investigated in prior ML GEO-based fire detection studies.

Despite these contributions, it is important to note some limitations of our study. The algorithm was primarily tested in regions of the continental United States and parts of North America, and only the CONUS and PACUS scan modes were used. The model was not tested on full-disk imagery or regions such as the Amazon or Patagonia, and therefore, continental-scale transferability remains unexplored. In addition, only the CatBoost model was evaluated, and it is possible that other advanced machine learning models could yield similar results.
Furthermore, the evaluation relied exclusively on VIIRS as a reference. Other LEO satellites with higher spatial resolution or in situ fire datasets were not used. Fire detections from these datasets are commonly included in the evaluation of GEO fire detection models \citep[e.g.,][]{HALL_2019_GOES_SEVIRI, Hall_2023_GOES_H8, LI_GOES_validation_20201}, as incorporating them can improve dataset labeling and evaluation.

Overall, this study shows that a CatBoost-based model can improve wildfire detection from GOES ABI imagery. It provides higher accuracy, lower false-alarm rates, and earlier detection than the GOES FDC product. These improvements highlight the potential of machine learning to strengthen operational wildfire monitoring and support more effective fire management and decision-making.

\section*{Funding}
This work was supported by the Israel Space Agency, which is part of the Ministry of Science and Technology, Israel, under Grant 3-18142; and the Jewish National Fund (JNF) under Grant 2006/22.

\section*{Author Contributions}
Conceptualization, A.V., B.N. and A.K.; methodology, A.V., B.N. and A.K.; software, A.V.; validation, A.V.; formal analysis, A.V.; investigation, A.V., B.N. and A.K.; resources, B.N. and A.K.; data curation, A.V.; writing---original draft preparation, A.V., B.N. and A.K.; writing---review and editing, A.V., B.N. and A.K.; visualization, A.V.; supervision, B.N. and A.K.; project administration, A.V.; funding acquisition, B.N. and A.K. All authors have read and agreed to the published version of the manuscript.

\section*{Disclosure statement}
The authors report there are no competing interests to declare.

\section*{Data availability statement}
The primary data that support the findings of this study are openly available in Zenodo at https://doi.org/10.5281/zenodo.19047285 \cite{asaf2026goes}. Additional data are available from the corresponding author, A.K., upon reasonable request.

\bibliographystyle{CSE-modified}
\bibliography{references}
\clearpage
\appendix
\renewcommand{\thesection}{\Alph{section}}
\renewcommand{\thetable}{\thesection.\arabic{table}}
\renewcommand{\thefigure}{\thesection.\arabic{figure}}
\counterwithin{table}{section}
\counterwithin{figure}{section}
\makeatletter
\renewcommand\@seccntformat[1]{Appendix~\csname the#1\endcsname\quad}
\makeatother
\section{Fire events}
\label{app:Historical Fire Events}
Table ~\ref{tab:fire_events} summarizes key attributes of each fire event, including the fire ID, name, the timestamp of the first VIIRS fire detection, U.S. state, centroid coordinates (latitude and longitude), and fire size (in km\textsuperscript{2}).

\begin{table}[H] 
\centering 
\caption{Summary of 51 historical fire events used for first-time detection comparison. The table lists the fire name, timestamp of first VIIRS detection, state, centroid coordinates (latitude and longitude) and fire size (in km\textsuperscript{2}).}
\vspace{0.5em}
\label{tab:fire_events} 
\resizebox{\textwidth}{!}{% 
\begin{filecontents*}{df_for_overleaf_no_size_class.csv}
ID,Fire event,VIIRS first detection time (UTC),State,Longitude,Latitude,Fire size (km2)
1,Area 6,28/1/2021 17:41,Florida,-81.268,27.682,17.5
2,Windmill,2/3/2021 19:06,Oklahoma,-97.265,34.398,189.2
3,CP-3,3/3/2021 18:41,Florida,-80.993,25.263,356.6
4,Dry Branch,5/3/2021 6:46,Alabama,-85.564,33.873,5.3
5,Pear Orchard,6/3/2021 17:46,Georgia,-83.41,34.537,9.6
6,Davis,7/3/2021 7:51,Mississippi,-89.176,31.128,9
7,Melder One,7/3/2021 19:11,Lousiana,-92.557,31.103,13.9
8,Narrows,7/3/2021 19:11,Georgia,-85.24,34.537,0.8
9,Mt Wiesner,7/3/2021 19:11,Alabama,-85.687,34.018,11
10,Blue Barrel,10/3/2021 18:11,Florida,-86.854,30.494,17.2
11,King,18/3/2021 7:46,Texas,-98.071,26.898,54.4
12,Addington,19/3/2021 18:46,Oklahoma,-97.917,34.293,15
13,Bentley,26/3/2021 8:36,Lousiana,-92.527,31.547,4.1
14,Candy Creek,28/3/2021 8:01,Oklahoma,-96.071,36.563,114.7
15,Blue Mountain,28/3/2021 19:16,Oklahoma,-94.837,34.835,14.4
16,Tour Road North,29/3/2021 19:01,Arkansas,-94.04,36.465,5.3
17,Oscar Ashton Road,17/5/2021 18:36,Florida,-81.522,29.89,3.6
18,Weeping Willow,19/5/2021 6:46,Florida,-81.972,27.703,8.9
19,Bobcat,29/5/2021 18:11,Florida,-80.737,26.384,155.2
20,Rare,11/6/2021 19:11,Florida,-82.381,30.374,18.1
21,Buck,28/10/2021 20:46,Texas,-99.405,32.986,107.6
22,Fall Branch,19/11/2021 18:51,Alabama,-85.792,33.466,4.4
23,Duck Nest,19/11/2021 18:51,Alabama,-85.822,33.49,2.3
24,Fodderstack,27/11/2021 8:21,Arkansas,-93.951,34.361,12
25,Tower Hills,5/12/2021 20:36,Oklahoma,-96.538,36.703,10.7
26,Cobb,15/12/2021 19:06,Oklahoma,-101.607,36.694,42.6
27,Walker,22/2/2024 19:21,Oklahoma,-94.517,34.827,11.9
28,Leaf,24/2/2024 18:46,Mississippi,-88.814,31.014,9.6
29,CMPT 270,24/2/2024 18:46,Mississippi,-89.577,32.11,1.3
30,Wired Grass,26/2/2024 8:31,Alabama,-86.159,33.183,0.5
31,Smokehouse Creek,26/2/2024 8:31,Texas,-100.666,35.837,6505
32,Zink,14/3/2024 8:11,Oklahoma,-96.28,36.32,50.6
33,Bell Spring,29/3/2024 6:51,Alabama,-85.892,33.523,5.3
34,Howell,4/4/2024 19:41,Oklahoma,-96.323,36.52,13.8
35,Hinson,5/4/2024 17:36,Florida,-81.508,29.069,12.5
36,Foote,12/6/2024 8:21,Arizona,-109.203,33.654,26.7
37,Sites,17/6/2024 21:31,California,-122.437,39.267,129.7
38,Blue Stem,1/7/2024 7:31,Texas,-95.276,29.113,12.9
39,Silver King,5/7/2024 20:51,Utah,-112.331,38.441,124.2
40,Deer Springs,7/7/2024 20:16,Utah,-112.267,37.241,75.3
41,Boise,9/8/2024 10:16,California,-123.473,41.248,94.2
42,Wiley Flat,3/9/2024 10:46,Oregon,-120.28,44.055,236.6
43,Boone,3/9/2024 20:31,California,-120.582,36.173,109.5
44,Line,6/9/2024 9:51,California,-117.066,34.16,265.9
45,Airport,8/9/2024 9:16,California,-117.493,33.662,137.9
46,Carson Six,16/10/2024 18:41,Mississippi,-89.055,30.541,7.4
47,Rush,25/10/2024 7:51,Oklahoma,-98.632,34.748,74.9
48,North Road,29/10/2024 19:36,Oklahoma,-96.341,36.629,23.9
49,Beech,2/12/2024 7:41,Oklahoma,-94.494,34.61,1.3
50,Middleton,16/2/2025 18:36,Texas,-94.395,29.619,16.6
51,Two Barrel,26/2/2025 18:46,Florida,-85.001,30.185,6.3
\end{filecontents*}
\csvreader[
    tabular=|>{\centering\arraybackslash}p{0.5cm}|c|c|c|c|c|c|c|,
    table head=\hline \centering ID & \begin{tabular}{@{}c@{}}Fire\\ event \end{tabular}& 
    \begin{tabular}{@{}c@{}}VIIRS first\\ detection time\\(UTC)\end{tabular} & 
    State & 
    Longitude & 
    Latitude & 
    \begin{tabular}{@{}c@{}}Fire size\\(km\textsuperscript{2})\end{tabular} 
   \\\hline,
    table foot=\hline
]{df_for_overleaf_no_size_class.csv}{}%
{\raggedright\csvcoli & \csvcolii & \csvcoliii & \csvcoliv & \csvcolv & \csvcolvi & \csvcolvii}
} 
\end{table}

\section{CatBoost Hyperparameters}
\label{app:CatBoost Model Appendix}
The CatBoost algorithm has several hyperparameters that control how the model learns. The \texttt{learning rate} sets how quickly the model adjusts its decision trees.
A smaller value, such as 0.01, makes learning more stable.
The number of \texttt{iterations} (500) determines how many trees the model builds. A larger number of iterations allow the model to capture more patterns in the data.

The \texttt{depth} of each tree (set to 8) controls how many decision levels the tree can create.
Deeper trees can represent more complex relationships between the input features.
The \texttt{L2 leaf regularization} value (set to 1) penalizes large leaf output values, preventing individual trees from excessively large prediction updates and thereby reducing overfitting.
The \texttt{bagging temperature} (set to 5) introduces randomness to the weights given to individual training samples, which also reduces overfitting.

To estimate these values, we performed a 5-fold cross-validation grid search.
We optimized for precision to reduce the false-alarm rate.

\begin{figure}[!ht]
    \centering
    \includegraphics[width=0.85\textwidth]{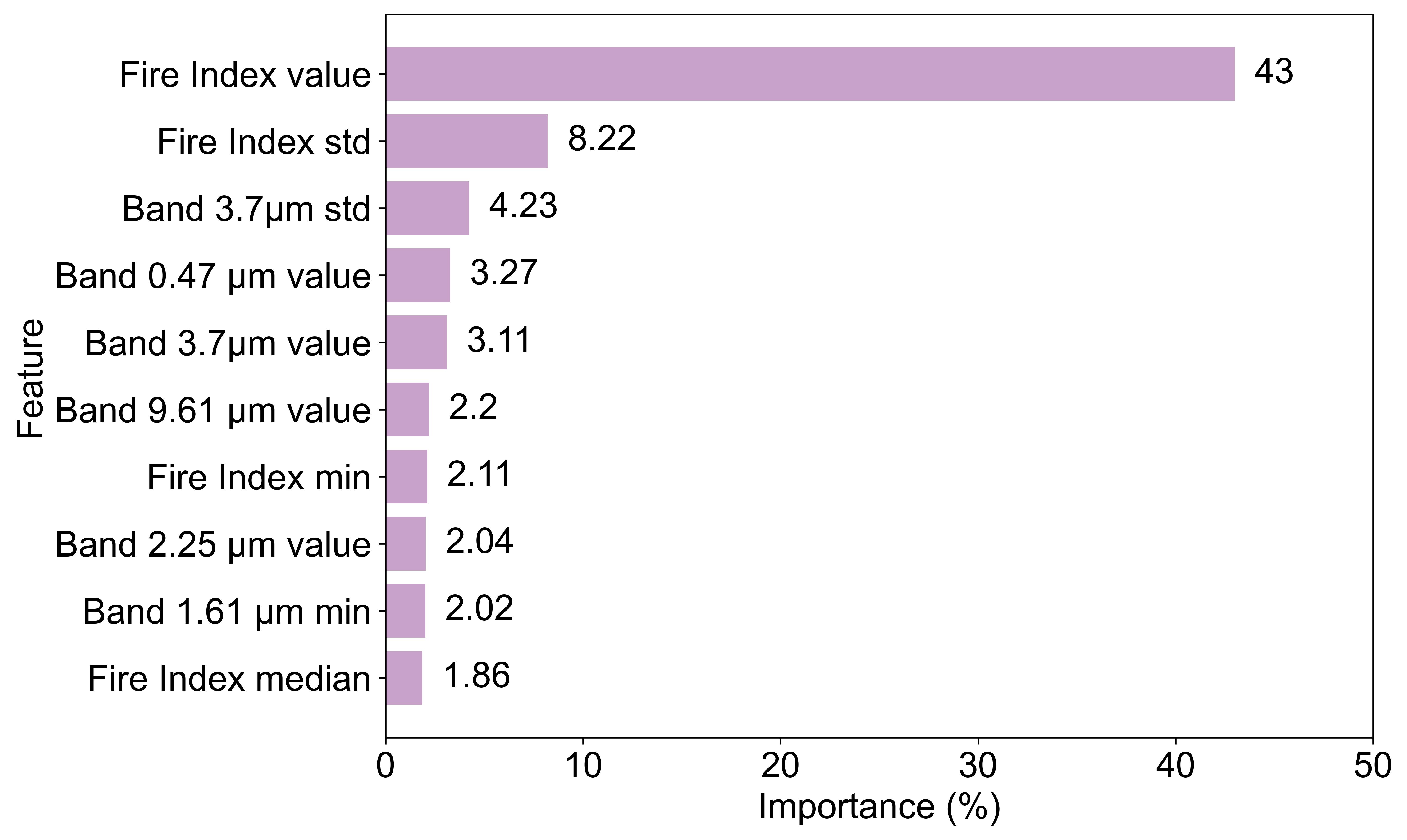}
    \caption{Ten most important features.}
    \label{fig:feature importance}
\end{figure}

Figure~\ref{fig:feature importance} shows the ten most important features in the final model. Several of the higher-ranked features come from the fire index and its statistics. 
This includes the fire index's standard deviation, which incorporates information from neighboring pixels and helps the model detect local variability. Additionally, the 3.7$\mu$m µm band, which is highly suitable for fire detection, demonstrated high importance. Interestingly, the 0.47$\mu$m visible band also received relatively high importance. It is possible that the algorithm uses this visible band to differentiate between daytime and nighttime conditions.
The figure highlights that both spectral information and simple contextual features contribute to the model’s predictions.

\end{document}